\documentclass{article}

\newif\ifarxivsty
\IfFileExists{arxiv.sty}{\arxivstytrue}{\arxivstyfalse}
\ifarxivsty
\usepackage{arxiv}
\else
\usepackage[margin=1in]{geometry}

\fi

\usepackage[utf8]{inputenc}
\usepackage[T1]{fontenc}
\IfFileExists{lmodern.sty}{\usepackage{lmodern}}{}
\usepackage{url}
\usepackage{booktabs}
\usepackage{amsfonts}
\usepackage{nicefrac}
\usepackage{microtype}
\usepackage{graphicx}
\usepackage{amsmath,amssymb}
\usepackage{caption}
\usepackage{subcaption}
\usepackage{ragged2e}
\usepackage{tabularx}
\usepackage{xcolor}
\usepackage{hyperref}
\usepackage{tikz}
\usetikzlibrary{positioning, arrows.meta}
\graphicspath{{./Figures/}{./images/}}

\AtBeginDocument{%
\setlength{\parindent}{0pt}%
\setlength{\parskip}{0.55\baselineskip plus 0.15\baselineskip minus 0.1\baselineskip}%
\setlength{\emergencystretch}{1.5em}%
\justifying}

\newcommand{\src}{\operatorname{src}}
\newcommand{\dst}{\operatorname{dst}}
\newcommand{\one}{\mathbf{1}}

\newcommand{\R}{\mathbb{R}}
\newcommand{\simplex}{\Delta}
\newcommand{\concat}{\mathbin{\Vert}}
\title{Conservative Hybrid Graph Networks for Process Systems with Learned Routing}

\author{%
Paolo Guida \\
Physical Science and Engineering \\
King Abdullah University of Science and Technology (KAUST) \\
Thuwal, Saudi Arabia \\
\texttt{paolo.guida@kaust.edu.sa} \\
}

\begin{document}
\maketitle
\setlength{\parindent}{0pt}

\begin{abstract}
Industrial process networks do not maintain a single effective topology while operating. Streams are throttled or bypassed, and units move between idle, transition and active regimes. Models of such systems are usually trained on measured state trajectories, while the operating mechanisms that generated those trajectories remain latent. An unconstrained graph network can fit such a trajectory without assigning a stable physical meaning to the recovered routing. This work addresses both problems through the Conservative Hybrid Graph Network (CHGN). The idea behind the architecture is to model routing, regime assignment, and removal rates using data-driven surrogates and add them to a fixed transport equation. The surrogate model allows extrapolation of dynamics learned on small process networks of $10$ to $20$ nodes to unseen, larger graphs of $25$ to $40$ nodes without retraining. While other Graph Neural Network (GNN)-based methods result in a $6\times10^{-2}$ to $9\times10^{-2}$ error, the CHGN algorithm presents an RMSE of $2.118\times10^{-3}$ under the same zero-shot protocol, with a gate MAE of $7.9\times10^{-3}$ and a regime accuracy of $94.3\%$.
On the fixed training topology, the corresponding values are $1.2\times10^{-2}$ and $96.4\%$.  The model is also tested on a real fluid-mixing pilot plant, where CHGN improves on persistence for held-out physical fault prediction while failing to predict manual operations (when governing valve actions are not observed). The benefits of the model can therefore be summarised as its ability to be used across various process topologies without retraining and to allow inspection of the underlying mechanisms that govern plants' decisions. 

\end{abstract}

\section{Introduction}
A structural affinity suggests modelling complex systems, such as chemical processes and industrial applications, as graphs. Recently, this has gained substantial interest \cite{torres2021and,leite2024modern,cortes2026phenomena}. Learned simulators on graphs usually assume a static topology \cite{battaglia2018relational} or a topology whose evolution is supplied to the model in advance \cite{sanchez2020gns,pfaff2021learning}. These assumptions do not completely represent industrial operation. In a plant, the effective connections between units change because streams are throttled or bypassed and because equipment is taken offline for safety, economic or maintenance reasons \cite{seborg2016process,alfares2022plant}. The measured state trajectory can remain available through sensor feedback, while the routing actions and operating regimes that generated it are only partially observed or not recorded at all \cite{kipf2018nri,linderman2017bayesian,seborg2016process}.

A network modelled with an unconstrained graph neural network can then experience two failure modes. The first concerns topological identifiability, referring to whether an observed trajectory assigns a unique meaning to a named routing variable inside the model. For instance, assume the stream associated with edge $e$ enters the state dynamics through the product $\hat w_e\phi_e(\mathbf x)$, where $\hat w_e$ is a learned routing weight and $\phi_e$ is a free learned message. If a positive constant $\alpha$ is introduced, the pair $(\alpha\hat w_e,\alpha^{-1}\phi_e)$ yields the same product and, therefore, the same state update. More generally, a sufficiently flexible message function can compensate for an incorrect routing weight. A small trajectory error, therefore, cannot uniquely identify an intended physical meaning.

While simple, the example above illustrates why a latent mechanism is not identifiable without additional structure \cite{schoelkopf2021toward}. Recovering an interaction graph requires a restriction on the decoder, a prior that breaks the symmetry, direct mechanism information, or a combination of these elements \cite{kipf2018nri}. The objective of the architecture proposed here is to remove the direct gate-message rescaling pathway by fixing the form through which routing enters transport.

A second problem is leakage. The latter consists of the artificial creation or loss of the conserved quantity during prediction. For an internal process network, the sum of all transport contributions should be zero because each internal stream leaves one unit and enters another. The total inventory can change only through explicit boundary feeds, removals or reactions. If a model predicts an increment $\Delta\mathbf x=f_\theta(\mathbf x)$ directly, and its parameterisation contains no corresponding balance restriction, nothing prevents material from appearing or vanishing over an autoregressive rollout. The model can still minimise a supervised prediction loss, but the resulting state evolution need not be compatible with the transport structure \cite{horie2024graph}.

This work intends to address both challenges. The proposed process-network formulation contains two mechanisms that can change independently: switch-controlled routing on the edges and discrete operating regimes at selected units. Internal transport is not represented by an unconstrained learned message. It is reconstructed from the signed incidence matrix, known transport-rate coefficients, the current-source inventory, and the predicted routing weights. The information supplied to the gate, regime, and removal heads is also separated, so that the named mechanisms cannot be exchanged freely within a single common decoder. Shared, permutation-equivariant mappings make the same model applicable to graphs with different numbers of units. The following sections describe the dynamical system and the software architecture and evaluate the model on controlled synthetic networks as well as on a physically distinct system built from real sensor data.
% ---------------------------------------------------------------------------
\section{The Conservative Hybrid Graph Network}
% ---------------------------------------------------------------------------
The proposed formulation combines ideas from latent graph inference, hybrid dynamical systems, learned simulation and structure-preserving modelling. The intended distinction is that existing methods usually address only part of the problem considered here.
Latent-structure methods infer fixed categorical edges \cite{kipf2018nri}, model changing graph content through memory \cite{kurenkov2023modeling}, and highlight the assumptions required for interpretable causal representations \cite{schoelkopf2021toward}; hybrid-system approaches represent global mode transitions \cite{poli2021nha,ferraritrecate2003pwa,bemporad2005,linderman2017bayesian}; graph simulators and neural operators approximate high-dimensional dynamics \cite{sanchez2020gns,pfaff2021learning,nabian24xmeshgraphnet,li2021fno,lu2021deeponet,kovachki2023neural,li2020neural}, including in continuous time \cite{chen2018neural}; structure-preserving models encode energetic constraints \cite{greydanus2019hamiltonian}, soft physical residuals \cite{raissi2019pinns}, local conservation and equivariance \cite{horie2024graph}, or discrete operator consistency \cite{li2024operator}, although such biases are not universally beneficial \cite{thangamuthu2022unravelling}; and process-system graph models address monitoring, soft sensing, fault diagnosis \cite{jia2025review,chen2021interaction,liu2024graph,brahmbhatt2024improved,seo2024graph,xu2025multiattention}, equipment design \cite{liu2025multi,wong2022graph}, and operating-phase switching \cite{teng2023machine,elmaz2023reinforcement,singer2021framework}, whereas CHGN performs forward simulation with exact internal transport balance by predicting state-dependent continuous edge routing and local unit regimes as distinct, potentially erroneous latent mechanisms rather than assuming a fixed known graph, a single global mode, or freely learned gate-conditioned messages.

\paragraph{Process graph and governing dynamics.}
The objective of the Conservative Hybrid Graph Network is to learn the evolution of a process in which stream routing and unit activity change during operation. The process is represented by a directed graph $G=(\mathcal V,\mathcal E)$ with $n=|\mathcal V|$ units and $m=|\mathcal E|$ internal streams. Its signed incidence matrix is
\begin{equation}
B\in\{-1,0,+1\}^{n\times m},\qquad
B_{ie}=
\begin{cases}
-1, & i=\src(e),\\
+1, & i=\dst(e),\\
0, & \text{otherwise}.
\end{cases}
\label{eq:incidence_matrix}
\end{equation}
Each internal edge, therefore, contributes once with a negative sign at its source and once with a positive sign at its destination.

The scalar state $x_i(t)\in\R_{\geq0}$ denotes the inventory of the tracked quantity in unit $i$. In the synthetic system, its evolution follows
\begin{equation}
\dot{x}_i(t)
=
\sum_{e\in\mathcal E}B_{ie}F_e(t)
+[E\mathbf u(t)]_i
-[\one_{\mathrm{sink}}]_i\,c_i(z_i(t))
\left[\kappa_i x_i(t)+\rho\eta_i x_i(t)^2\right].
\label{eq:gt_nodewise}
\end{equation}
The first term is the net internal transport entering or leaving unit $i$. The second term represents external feeds. In the latter, $E\in\R^{n\times p}$ is a known feed-incidence matrix and $\mathbf u(t)\in\R^p$ contains the imposed boundary inputs. The final term represents material removal or consumption at the nodes selected by the static mask $\one_{\mathrm{sink}}\in\{0,1\}^n$. The coefficient $\kappa_i$ defines a first-order contribution, $\eta_i$ defines a quadratic contribution and $\rho$ scales the latter globally. The multiplier $c_i(z_i)$ adjusts these rates based on the local operating regime. For nodes without a regime variable, this multiplier is fixed.

Internal flow is not arbitrary. For each edge,
\begin{equation}
F_e(t)=q_e\,w_e(t)\,x_{\src(e)}(t),
\qquad q_e\geq0,
\label{eq:true_edge_flux}
\end{equation}
where $q_e$ is a known transport-rate coefficient with units of inverse time and $w_e(t)$ is the routing weight of that edge. Routing occurs through $S$ switches. Switch $s$ acts on two outgoing edges, denoted $e_s^1$ and $e_s^2$, and is governed by
\begin{equation}
g_s(t)=\sigma\!\left(\beta_s\left[\varphi_s(\mathbf x(t),\mathbf u(t))-\theta_s^g\right]\right),
\qquad
\sigma(v)=\frac{1}{1+e^{-v}},
\label{eq:routing_sigmoid}
\end{equation}
with
\begin{equation}
w_{e_s^1}(t)=g_s(t),\qquad
w_{e_s^2}(t)=1-g_s(t),\qquad
w_e(t)=1\quad\text{for unswitched edges}.
\label{eq:true_routing_weights}
\end{equation}
The function $\varphi_s$ combines the current state and input, $\theta_s^g$ shifts the switching threshold and $\beta_s>0$ controls the steepness of the response. The sigmoid produces a soft routing weight that approaches zero or one as the input moves away from the threshold.

A subset of units $\mathcal M\subseteq\mathcal V$, with $M=|\mathcal M|$, also experiences local regime changes. Each unit $j\in\mathcal M$ can be idle, transitioning or active, represented by $z_j(t)\in\{0,1,2\}$. The simulator assigns these states through a local threshold rule parameterised by $\theta_j^z$. The conditioning vector
\begin{equation}
\boldsymbol\mu=(\boldsymbol\theta^g,\boldsymbol\theta^z,\rho)
\label{eq:conditioning_vector}
\end{equation}
contains the switch thresholds, the local regime thresholds and the global quadratic scale. These quantities are supplied as operating conditions. The instantaneous gates $g_s(t)$ and regimes $z_j(t)$ are not supplied to the state predictor and must be recovered from the observed history.

\paragraph{Algorithm architecture.}
Rather than replacing the complete vector field with a black-box neural
operator, CHGN learns four modules,
\begin{equation*}
\{\Phi_{\theta_\Phi},\Gamma_{\theta_\Gamma},
Z_{\theta_Z},R_{\theta_R}\},
\qquad
\boldsymbol\Theta
=
\{\theta_\Phi,\theta_\Gamma,\theta_Z,\theta_R\},
\end{equation*}
where $\boldsymbol\Theta$ denotes their trainable parameters.
The first module, $\Phi_{\theta_\Phi}$, is a shared spatial GNN that maps
the state history and unit information into a $d$-dimensional latent
space. At time index $\tau$, the input associated with node $i$ is
\begin{equation}
\mathbf a_{i,\tau}
=
H_{\tau,i,:}^{\top}
\concat
\mathbf e(\mathrm{type}_i)
\concat
\mathbf u_\tau
\in\R^{h+d_e+p},
\label{eq:node_features}
\end{equation}
where $\concat$ denotes concatenation, $\mathbf e(\mathrm{type}_i)\in\R^{d_e}$ is a learned unit-type embedding, $\mathbf u_\tau\in\R^p$ collects the external inputs and
\begin{equation}
H_\tau
=
[\mathbf x_{\tau-h+1},\ldots,\mathbf x_\tau]
\in\R^{n\times h}
\label{eq:history_matrix}
\end{equation}
is the variable-history matrix. The complete node-feature matrix is
\begin{equation}
A_\tau
=
H_\tau
\concat
\mathcal T
\concat
\one_n\mathbf u_\tau^\top
\in\R^{n\times(h+d_e+p)},
\label{eq:node_feature_matrix}
\end{equation}
where the $i$-th row of $\mathcal T\in\R^{n\times d_e}$ is
$\mathbf e(\mathrm{type}_i)^\top$. The spatial encoder then gives
\begin{equation}
\boldsymbol\nu_{i,\tau}^{\top}
=
\left[
\Phi_{\theta_\Phi}(A_\tau,G,\mathbf q)
\right]_{i,:},
\qquad
\boldsymbol\nu_{i,\tau}\in\R^d,
\label{eq:node_encoder_head}
\end{equation}
where $G$ is the static process graph and $\mathbf q$ contains the static edge features, including the transport-rate coefficients $q_e$. A permutation-invariant pooled vector $\bar{\boldsymbol\nu}_\tau$ provides global context without fixing the number or ordering of nodes.

The second module is the gate head $\Gamma_{\theta_\Gamma}$. Each of the $S$ routers is indexed by $s$; router $s$ is located at node $v_s$ and splits its throughput between destination nodes $d_s^1$ and $d_s^2$ along edges $e_s^1$ and $e_s^2$. It predicts the soft routing variable
\begin{equation}
\hat g_{s,\tau}
=
\Gamma_{\theta_\Gamma}\!\left(
\boldsymbol\nu_{v_s,\tau},
\boldsymbol\nu_{d_s^1,\tau},
\boldsymbol\nu_{d_s^2,\tau},
\bar{\boldsymbol\nu}_\tau,
\mathrm{switch}_s,
\mathbf u_\tau,
\boldsymbol\xi_s^g
\right),
\qquad
\hat g_{s,\tau}\in(0,1),
\label{eq:gate_head_prediction}
\end{equation}
where $\mathrm{switch}_s$ identifies the switch type and $\boldsymbol\xi_s^g$ contains its static physical parameters, such as the switching threshold $\theta_s^g$. Neither contains the realised routing target. The predicted gate induces complementary edge weights,
\begin{equation}
\hat w_{e_s^1,\tau}
=
\hat g_{s,\tau},
\qquad
\hat w_{e_s^2,\tau}
=
1-\hat g_{s,\tau},
\qquad
\hat w_{e,\tau}
=
1
\quad\text{for unswitched edges}.
\label{eq:gated_routing_weights}
\end{equation}
The two branches share the same transport-rate coefficient, $q_{e_s^1}=q_{e_s^2}=q_s$, so their combined outflow is independent of the gate:
\begin{equation}
q_s\hat g_{s,\tau}\hat x_{v_s,\tau}
+q_s(1-\hat g_{s,\tau})\hat x_{v_s,\tau}
=
q_s\hat x_{v_s,\tau}.
\label{eq:router_flow_split}
\end{equation}
The role of the GNN is therefore to provide state- and topology-dependent features to the gate head.
The third module is the regime head $Z_{\theta_Z}$ acting locally on every unit that allows for the occurrence of different regimes $j\in\mathcal M$ and produces
\begin{equation}
\boldsymbol\ell_{j,\tau}
=
Z_{\theta_Z}\!\left(
H_{\tau,j,:}^{\top},
\mathbf e(\mathrm{type}_j),
\mathbf u_\tau,
\boldsymbol\xi_j^z
\right),
\qquad
\hat{\boldsymbol\pi}_{j,\tau}
=
\operatorname{softmax}(\boldsymbol\ell_{j,\tau})
\in\simplex^2,
\label{eq:regime_head_prediction}
\end{equation}
where $\boldsymbol\xi_j^z$ contains the static regime parameters, including the threshold $\theta_j^z$, and $\hat{\boldsymbol\pi}_{j,\tau}$ contains the probabilities of the idle, transition and active regimes. The final module is the sink head $R_{\theta_R}$. It predicts a bounded, state-dependent removal-rate coefficient at every designated sink,
\begin{equation}
r_{i,\tau}
=
r_{\max}\sigma\!\left(
R_{\theta_R}\!\left(
H_{\tau,i,:}^{\top},
\mathbf e(\mathrm{type}_i),
\mathbf u_\tau,
\rho
\right)
\right),
\label{eq:sink_rate_prediction}
\end{equation}
where $\sigma(\cdot)$ is the logistic function of Equation~\eqref{eq:routing_sigmoid}. The corresponding removal term is assembled algebraically as
\begin{equation}
s_{i,\tau}
=
r_{i,\tau}\hat x_{i,\tau}
[\one_{\mathrm{sink}}]_i
\bar c_{i,\tau}.
\label{eq:material_removal_term}
\end{equation}
This parameterisation admits the first- plus second-order synthetic removal whenever the target coefficient
\begin{equation}
r^\star_{i,\tau}
=
\kappa_i+\rho\eta_i\hat x_{i,\tau}
\label{eq:target_removal_rate}
\end{equation}
lies inside the representable interval $(0,r_{\max})$. Pointwise, the required sink-head pre-activation is
\begin{equation}
R^\star_{i,\tau}
=
\log\!\left(
\frac{r^\star_{i,\tau}}
{r_{\max}-r^\star_{i,\tau}}
\right).
\label{eq:sink_inverse_mapping}
\end{equation}
The neural sink head can approximate this mapping over the bounded state domain used for training, while the endpoints can only be approached asymptotically. The bounded parameterisation prevents the learned removal rate from becoming negative or unbounded and makes its representability condition explicit.

\begin{figure}[t]
\centering
\includegraphics[width=\linewidth]{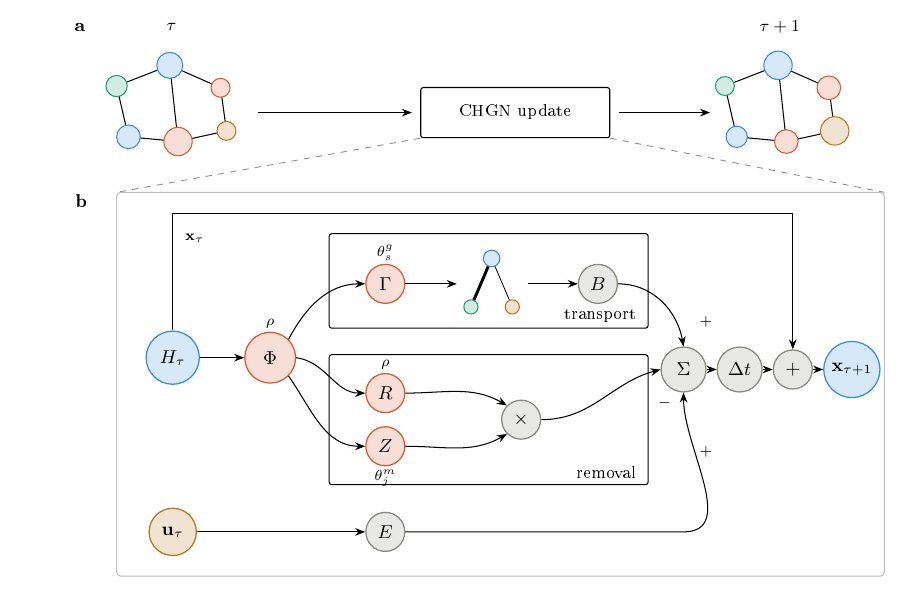}
\caption{\textbf{One-step CHGN update.} The spatial encoder maps the state history, unit types, external inputs and static process graph into topology-dependent node representations used by the routing head. Local
heads infer the unit regimes and bounded removal rates from the corresponding state histories and unit information. The predicted quantities are then inserted into the fixed incidence-based transport law and advanced by one time step.}
\label{fig:chgn-architecture}
\end{figure}

\paragraph{Assembly.}
Once the four modules have produced the routing weights, the regime
multipliers and removal terms, the future state is obtained by reconstructing the process balance. Throughout the rollout, the state entering the right-hand side is the model's own estimate, with the true initial history supplied at the beginning of the rollout. The predicted material flow on edge $e$ is
\begin{equation}
\hat F_{e,\tau}
=
q_e\hat w_{e,\tau}\hat x_{\src(e),\tau}.
\label{eq:predicted_edge_flux}
\end{equation}
The complete derivative then reads
\begin{equation}
\dot{\hat{\mathbf x}}_\tau
=
B\hat{\mathbf F}_\tau
+E\mathbf u_\tau
-\mathbf s_\tau.
\label{eq:state_update}
\end{equation}
An explicit Euler step advances the state,
\begin{equation}
\tilde{\mathbf x}_{\tau+1}
=
\hat{\mathbf x}_\tau
+\Delta t\,\dot{\hat{\mathbf x}}_\tau,
\qquad
\hat{\mathbf x}_{\tau+1}
=
\max(\mathbf 0,\tilde{\mathbf x}_{\tau+1}),
\label{eq:explicit_euler_loop}
\end{equation}
where the maximum is applied element-wise as a numerical non-negativity safeguard. This safeguard would alter the balance if it were activated. In the reported experiments, its correction is identically zero at every evaluated rollout step, so the unmodified Euler update governs the rollout.

The internal transport term is conservative by construction. Every column of $B$ contains one $-1$ and one $+1$, so
$\one_n^\top B=\mathbf 0^\top$ and therefore
\begin{equation}
\one_n^\top\dot{\hat{\mathbf x}}_\tau
=
\underbrace{\one_n^\top B\hat{\mathbf F}_\tau}_{=0}
+\one_n^\top E\mathbf u_\tau
-\one_n^\top\mathbf s_\tau.
\label{eq:mass_balance}
\end{equation}
The identity guarantees that routing can only redistribute material between units at a given instant. After each step, the history matrix is updated to include $\hat{\mathbf x}_{\tau+1}$, and the calculation repeats in an autoregressive manner.

\paragraph{Training.}
The four modules are trained jointly by unrolling
Equation~\eqref{eq:explicit_euler_loop} for $T$ transitions from a true
initial history ending at $\tau=0$. The transition is evaluated at time
$\tau$ predicts $\mathbf x_{\tau+1}$, with
$\tau=0,\ldots,T-1$. The objective combines state prediction with weak
supervision of the routing and regime mechanisms,
\begin{equation}
\mathcal L
=
\mathcal L_{\mathrm{state}}
+\lambda_g\mathcal L_{\mathrm{gate}}
+\lambda_z\mathcal L_{\mathrm{regime}}.
\label{eq:joint_loss}
\end{equation}
The term is
\begin{equation}
\mathcal L_{\mathrm{state}}
=
\sum_{\tau=0}^{T-1}\omega_{\tau+1}
\frac{1}{n}\sum_{i=1}^{n}
\left(
\frac{\hat x_{i,\tau+1}-x_{i,\tau+1}}{\alpha_i}
\right)^2,
\label{eq:state_loss}
\end{equation}
where $\alpha_i>0$ is the state-normalisation constant and
$\omega_{\tau+1}$ weights the forecast horizon.

The gate loss is a binary cross-entropy applied to the supervised soft gate targets $g_{s,\tau}\in[0,1]$. In contrast, the regime loss is a categorical cross-entropy applied to the supervised regime labels $z_{j,\tau}$. Let $m^g_{s,\tau}\in\{0,1\}$ and $m^z_{j,\tau}\in\{0,1\}$ indicate the labelled gate and regime entries, respectively. The losses are
\begin{equation}
\begin{aligned}
\mathcal L_{\mathrm{gate}}
&=
\frac{1}{N_g}
\sum_{\tau=0}^{T-1}\sum_{s=1}^{S}
m^g_{s,\tau}
\operatorname{BCE}(\hat g_{s,\tau},g_{s,\tau}),\\
\mathcal L_{\mathrm{regime}}
&=
\frac{1}{N_z}
\sum_{\tau=0}^{T-1}\sum_{j\in\mathcal M}
m^z_{j,\tau}
\operatorname{CE}(\boldsymbol\ell_{j,\tau},z_{j,\tau}).
\end{aligned}
\label{eq:mechanism_losses}
\end{equation}
where
\begin{equation}
N_g
=
\sum_{\tau=0}^{T-1}\sum_{s=1}^{S}m^g_{s,\tau},
\qquad
N_z
=
\sum_{\tau=0}^{T-1}\sum_{j\in\mathcal M}m^z_{j,\tau}.
\label{eq:mechanism_label_counts}
\end{equation}
Thus, both terms are averaged over all labelled entries. The sink head receives no direct supervision and is constrained only through the unrolled state-prediction loss. Optimisation uses AdamW~\cite{loshchilov2017decoupled} with a learning rate of $2\times10^{-3}$, a weight decay of $10^{-6}$, over $150$ epochs and gradient-norm clipping at $1.0$. The history window is $h=5$ steps, the latent width is
$d=96$, and the encoder uses three message-passing rounds. 

% ---------------------------------------------------------------------------
\section{Results}\label{sec:results}
% ---------------------------------------------------------------------------

\paragraph{Systems and organisation of the evaluation.}
The proposed architecture is evaluated on two systems. The first is a family of synthetic process networks in which the states, gates, regimes, interventions and balance laws are all known. This system allows trajectory prediction to be separated from mechanism recovery, and permits transfer, counterfactual response, and conservation to be examined independently of one another. The second test is a real fluid-mixing pilot plant, in which the same architectural principles are evaluated across operating conditions and anomaly categories not included in training.

The first test evaluates whether the learned process graph can be extrapolated to larger systems without retraining. The second evaluates whether naming the mechanisms is necessary or whether similar accuracy can be achieved without that information. The third evaluates performance when predicting system dynamics is the only objective; it highlights one limitation of the proposed algorithm in its current configuration.

\subsection{Experiment 1: Synthetic process-network benchmark}
\label{sec:experiment-synthetic}

\paragraph{Transfer to unseen graphs.}
\label{sec:transfer}
CHGN is trained on a distribution of $10$ to $20$-node process graphs and then evaluated, without fine-tuning, on eight unseen $25$ to $40$-node graphs. The protocol contains $64$ graph trajectory evaluations per seed and five independently trained seeds. Two shared-weight baselines are trained on the same multi-topology distribution and evaluated under the same zero-shot conditions: a Shared Dynamic GNN and a Shared Conservative GNN\@. The latter uses a conservative aggregation but does not include the complete CHGN factorisation into fixed physical transport, routing gates, and local regime variables. The comparison is therefore matched on parameter sharing and on the training distribution, and differs only in the internal factorisation. 

The difference is consistent across both graphs and seeds. CHGN remains near $2\times10^{-3}$ state RMSE over the unseen topologies, while the two shared baselines remain approximately in the $6\times10^{-2}$--$9\times10^{-2}$ range. CHGN also keeps gate MAE below $10^{-2}$ and maintains regime accuracy at approximately $94\%$. The separation persists as graph size increases from 25 to 40 nodes, as shown in Figure~\ref{fig:transfer} and, in more detail, in Appendix~\ref{app:transfer}. The transfer advantage, therefore, cannot be attributed solely to graph-size-independent parameter sharing. The weaker performance of the Shared Conservative GNN also shows that conservation alone does not reproduce the complete result. A secondary comparison against baselines retrained separately on each held-out graph is reported in Appendix~\ref{app:transfer}; that tests graph-specific fitting.

\begin{figure}[t]
\centering
\includegraphics[width=\textwidth]{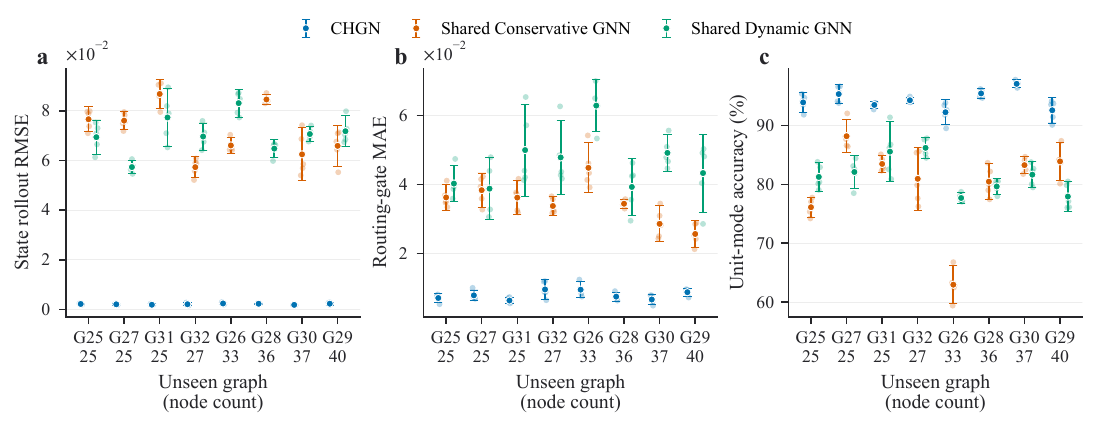}
\caption{\textbf{Zero-shot transfer across unseen process graphs.} CHGN and two shared-weight baselines are trained on the same $10$--$20$-node graph distribution and evaluated without retraining on unseen $25$--$40$-node topologies. Points and bars show the mean and $95\%$ Student-$t$ interval across seeds; faint points are individual-seed means. Node count orders graphs.}
\label{fig:transfer}
\end{figure}

\begin{figure}[t]
\centering
\includegraphics[width=\textwidth]{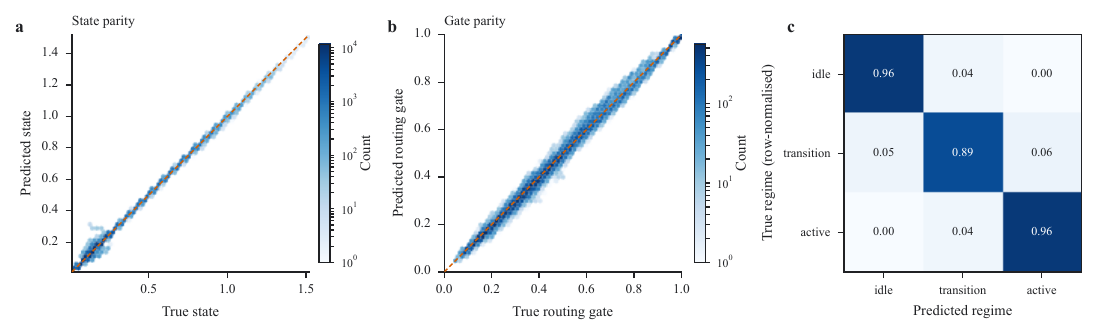}
\caption{\textbf{Calibration of the three predicted quantities.} Predicted versus true (a) inventory and (b) routing gate, shown as hexagonally binned log-count densities; the dashed line is parity. (c) Row-normalised regime confusion matrix. Gates are calibrated across $(0,1)$, and regime errors occur only between adjacent classes.}
\label{fig:parity}
\end{figure}

\paragraph{Recovery and calibration of the named mechanisms.}
\label{sec:calibration}
The transfer result raises the question of what the model has actually recovered. Figure~\ref{fig:parity} examines the three predicted quantities on the training topology. State predictions follow parity across the full inventory range, with most of the dispersion occurring below an inventory of $0.2$. Gate predictions cover the complete interval $(0,1)$, including values close to $0.5$. The absence of a gap around the switching threshold shows that the model has not reduced the task to a binary classification of the active branch. Regime accuracy is $0.96$ for idle and active and $0.89$ for transition. All regime errors occur between adjacent classes, consistent with errors near regime boundaries rather than with collapse to a single dominant class. The corresponding full-trajectory rollout and a unit-by-unit audit on the flowsheet are reported in Appendix~\ref{app:reconstruction}.

\begin{figure}[t]
\centering
\includegraphics[width=\textwidth]{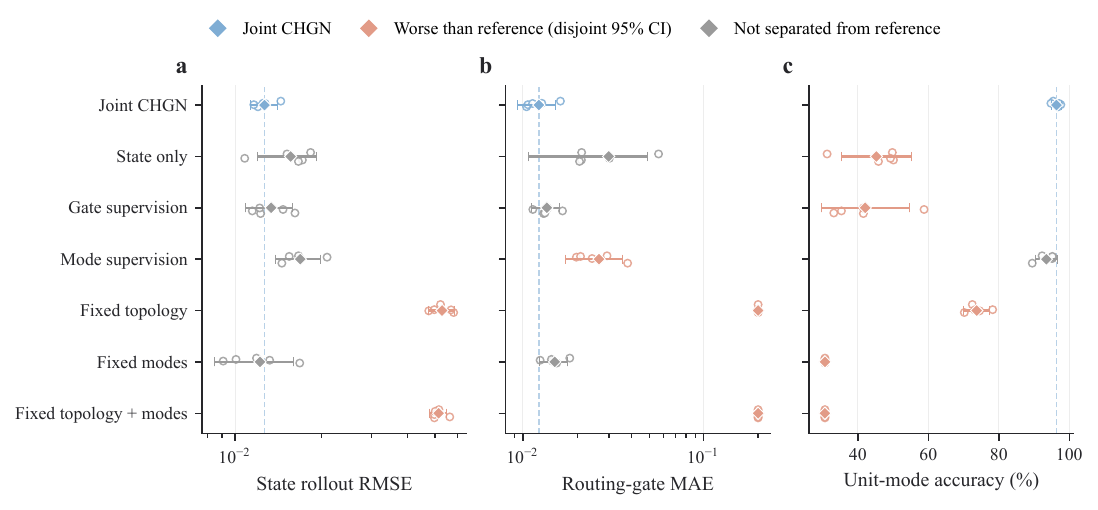}
\caption{\textbf{Mechanism ablations, five seeds per variant.} Circles show seeds; diamonds and bars show the mean and two-sided $95\%$ Student-$t$ interval. The dashed line marks the joint-model mean. Red indicates an interval separated from the reference in the worst direction; grey indicates no separation.}
\label{fig:ablations}
\end{figure}

\paragraph{Mechanism ablations.}
\label{sec:ablations}
Table~\ref{tab:ablations} and Figure~\ref{fig:ablations} separate state prediction from the recovery of the named mechanisms. Routing is obviously extremely relevant in the forumlation.  Fixing the routing mechanism increases state RMSE from $0.01267$ to $0.05293$. Across five matched seeds, the mean increase is $0.0403$, and the paired bootstrap $95\%$ interval $[0.0362,0.0443]$ excludes zero. Fixing the regime mechanism has a different effect. Regime accuracy falls to $30.7\%$, but the mean state-RMSE change is only $-4.6\times10^{-4}$, with a paired $95\%$ interval of $[-2.69\times10^{-3},2.60\times10^{-3}]$. The aggregate state error is therefore not separated from the joint model.

Direct mechanism supervision is nevertheless necessary to recover the variables by name. Without regime labels, regime accuracy falls from $96.4\%$ to $45.3\%$ for the state-only variant and to $42.1\%$ for the gate-supervised variant, while the state error changes much less. Conversely, the regime-supervised model retains $93.5\%$ regime accuracy, but its gate MAE increases to $0.0264$. The two heads, therefore, receive distinct supervisory information. Routing has the dominant effect on the aggregate state in this benchmark. At the same time, the regime variable can be recovered accurately even though its contribution to the total state error is smaller.

\begin{table}[t]
\centering
\small
\caption{Mechanism ablations. Mean $\pm$ s.d.\ across five independently trained seeds. Entries without intervals are clamped and constant across seeds.}
\label{tab:ablations}
\begin{tabular*}{\textwidth}{@{\extracolsep{\fill}}lccc@{}}
\toprule
Variant & State RMSE $\downarrow$ & Gate MAE $\downarrow$ & Regime acc. $\uparrow$ \\
\midrule
Joint CHGN              & $0.01267 \pm 0.00108$ & $0.01225 \pm 0.00234$ & $96.37 \pm 1.18\%$ \\
State only              & $0.01564 \pm 0.00294$ & $0.02985 \pm 0.01535$ & $45.26 \pm 8.00\%$ \\
Gate supervision        & $0.01337 \pm 0.00199$ & $0.01357 \pm 0.00189$ & $42.08 \pm 10.06\%$ \\
Regime supervision      & $0.01690 \pm 0.00246$ & $0.02639 \pm 0.00745$ & $93.51 \pm 2.58\%$ \\
Fixed routing           & $0.05293 \pm 0.00454$ & $0.20051$             & $73.74 \pm 2.97\%$ \\
Fixed regimes           & $0.01221 \pm 0.00303$ & $0.01499 \pm 0.00208$ & $30.66\%$ \\
Fixed routing + regimes & $0.05151 \pm 0.00281$ & $0.20051$             & $30.66\%$ \\
\bottomrule
\end{tabular*}
\end{table}

\paragraph{Interventional and counterfactual response.}
\label{sec:intervention-summary}
An ablation establishes if a mechanism is structurally relevant, but not that it responds correctly when it is manipulated. Three audits address the latter and are reported in full in Appendix~\ref{app:intervention}. Substituting mechanism variables into the same trained operator increases the one-step all-unit RMSE by a factor of $4.7$ when routing is held fixed and by a factor of $4.8$ in the worst ablation, while replacing the learned mechanisms with their true values leaves the error essentially unchanged. Sweeping each switch threshold over values absent from the held-out trajectory reproduces the true logistic response across the complete range, with a gate MAE between $9.2\times10^{-3}$ and $2.1\times10^{-2}$ and a monotone fraction of $1.000$. Propagating a threshold change through a $40$-step rollout without teacher forcing gives an intervention-effect RMSE of $6.4\times10^{-4}$ over $240$ routing interventions and $8.0\times10^{-5}$ over $192$ regime interventions. The recovered gates, therefore, retain their role under manipulation, and not only on the factual trajectory from which they were inferred.

\paragraph{Fixed-graph benchmark.}
\label{sec:baselines}
Table~\ref{tab:baselines} shows the performance of the proposed solver against other methods. Seven models are compared on a single fixed 20-unit graph, and CHGN does not achieve the lowest point-prediction error in this setting. The LSTM reaches a state RMSE of $0.00520\pm0.00035$, compared with $0.01267\pm0.00108$ for CHGN, and is therefore approximately $2.4\times$ more accurate. It also reaches the lowest gate MAE and the lowest balance-model discrepancy.
 The balance-model discrepancy measures the agreement between the predicted state evolution and the complete known inventory budget and therefore includes state, boundary, and removal errors. The structural residual $\one^\top B\hat{\mathbf F}$ instead isolates internal transport and is identically zero in exact arithmetic for CHGN\@. A flexible sequence model can consequently obtain a lower balance-model discrepancy through a more accurate trajectory fit without carrying an explicit internal conservation guarantee.

The fixed-graph comparison, therefore, clarifies the intended scope of the formulation. Its structural restrictions do not guarantee the best fit when the topology does not change, and state prediction is the only objective. Their purpose is to preserve a defined transport law, expose named mechanisms that can be intervened on and provide graph-local mappings that remain applicable when the process topology changes.

\begin{table}[t]
\centering
\small
\caption{Fixed-graph benchmark, five seeds per model. Bold marks the best mean, not statistical separation. Balance discrepancy measures agreement with the known full balance model and is distinct from the internal transport residual $\one^\top B\hat{\mathbf F}$.}
\label{tab:baselines}
\begin{tabular*}{\textwidth}{@{\extracolsep{\fill}}lcccc@{}}
\toprule
Model & State RMSE $\downarrow$ & Gate MAE $\downarrow$ & Regime acc. $\uparrow$ & Balance disc. $\downarrow$\\
\midrule
CHGN          & $0.01267\pm0.00108$ & $0.01225\pm0.00234$ & $96.37\pm1.18\%$ & $0.01983\pm0.00394$\\
LSTM          & $\mathbf{0.00520\pm0.00035}$ & $\mathbf{0.00621\pm0.00037}$ & $94.34\pm0.22\%$ & $\mathbf{0.01091\pm0.00082}$\\
MLP           & $0.01630\pm0.00131$ & $0.01661\pm0.00218$ & $92.97\pm0.31\%$ & $0.01574\pm0.00121$\\
Dynamic GNN   & $0.03421\pm0.00035$ & $0.01072\pm0.00020$ & $\mathbf{96.57\pm0.11\%}$ & $0.11376\pm0.00157$\\
NHA-inspired  & $0.03437\pm0.00037$ & $0.01061\pm0.00028$ & $96.35\pm0.12\%$ & $0.11430\pm0.00133$\\
NRI-inspired  & $0.05192\pm0.00032$ & $0.05263\pm0.00378$ & $91.64\pm0.63\%$ & $0.11667\pm0.00220$\\
Fixed GNN     & $0.06607\pm0.00382$ & $0.20051$           & $81.57\pm3.14\%$ & $0.22629\pm0.02543$\\
\bottomrule
\end{tabular*}
\end{table}

\paragraph{Structural consistency.}
\label{sec:numerics-summary}
A numerical audit separates approximation error from structural consistency and is reported in full in Appendix~\ref{app:numerics}. Across independently trained seeds, the state-derivative RMSE is approximately $9.8\times10^{-3}$, while the maximum internal-transport residual $\max_\tau|\one^\top B\hat{\mathbf F}_\tau|$ is approximately $1.2\times10^{-7}$ and is therefore at the level expected from floating-point accumulation. The non-negativity safeguard in Equation~\eqref{eq:explicit_euler_loop} is inactive over every entry and rollout step. The remaining derivative error, therefore, reflects the approximation of the non-transport dynamics and the difference between the Euler update and the RK4 integration used to generate the synthetic trajectories, rather than a violation of the incidence-based identity.

\subsection{Experiment 2: Real fluid-mixing system}
\label{sec:fluidmix}

The second experiment applies the architecture to a physically different, real fluid-mixing pilot plant with four tanks, two pumps and eleven valves \cite{ramonat2025fluidmix}. The process graph contains four tanks, eight pipe junctions, three external ports and eighteen edges. In this case, the graph is constructed from the P\&ID naming convention rather than inferred from the data.

This experiment transfers the architectural constraints, not the parameters learned on the synthetic networks. A new CHGN model is trained on the fluid-mixing data while retaining the graph-based conservative accumulation and the separation between observed state, external forcing and mechanism-specific quantities. The mappings required by the physical system are learned from that dataset. 

\begin{figure}[htbp]
\centering
\includegraphics[width=\textwidth]{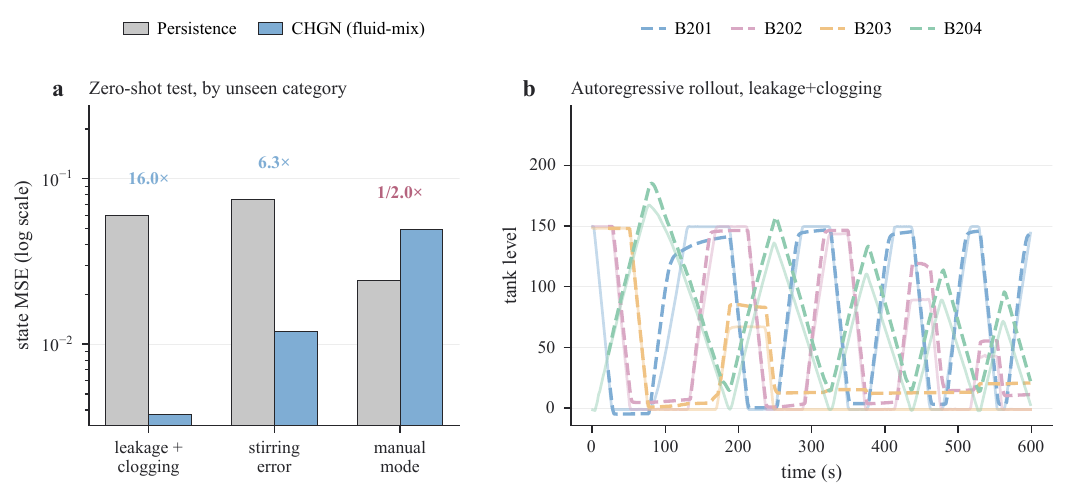}
\caption{\textbf{Structure-preserving architecture on held-out fluid-mixing conditions.} (a) State MSE for persistence and CHGN on three anomaly categories held out from training; leakage+clogging and stirring error each contain one session, while manual mode contains eight. (b) Autoregressive rollout on the leakage+clogging session. }
\label{fig:fluidmix}
\end{figure}

Compared with persistence, CHGN reduces state MSE by a factor of $16.0$ on the held-out leakage-plus-clogging session and by a factor of $6.3$ on the stirring-error session. These values correspond to improvements of $4.0\times$ and $2.5\times$ in RMSE\@. The autoregressive rollout in Figure~\ref{fig:fluidmix} follows the cyclic fill--drain pattern of the plant. Manual mode produces the opposite result. CHGN is $2.0\times$ worse than persistence in MSE and $1.4\times$ worse in RMSE, because the manual valve actions that determine the subsequent trajectory are not included in the model inputs. The experiment, therefore, separates faults whose dynamics can be inferred from the observed variables from operating changes driven by an unobserved exogenous control.

% ---------------------------------------------------------------------------
\section{Discussion and limitations}\label{sec:limitations}
% ---------------------------------------------------------------------------

The two experiments support the following conclusions. On synthetic networks, it was demonstrated that CHGN can be evaluated without retraining on larger process graphs, recover named routing and regime variables, respond consistently when their thresholds are changed, and preserve the internal transport balance by construction. These properties, however, do not improve trajectory prediction.

The matched zero-shot transfer comparison narrows the possible explanations for the principal result. CHGN, the Shared Dynamic GNN and the Shared Conservative GNN all use graph-size-independent parameters and are trained on the same $10$ to $20$-node distribution. Only CHGN retains low state and mechanism errors on the unseen $ 25$- to $40$-node graphs. The result cannot, therefore, be attributed only to shared message-passing weights. The weaker Shared Conservative GNN also shows that conservation alone is insufficient. The current experiments support the combined role of the fixed transport law, named routing variables, information partitioning and graph-local parameter sharing. Isolating the individual contribution of each ingredient is left to future work.

The mechanism results in separate recoverability from dynamical importance. Routing is strongly load-bearing: fixing the routing variable produces a five-seed increase in state error whose paired bootstrap interval excludes zero, and routing interventions propagate through the closed-loop trajectory. The regime variable is recovered accurately but has a much smaller effect on aggregate state error in the principal benchmark. Both variables also receive weak direct supervision during training. The experiments therefore show that the architecture can assign a stable role to supervised mechanism variables and preserve that role during rollout. 

A related limitation concerns the meaning of identifiability itself. Fixing the transport law to $\hat F_e=q_e\hat w_e\hat x_{\src(e)}$ removes the direct reciprocal scaling between a routing weight and an arbitrary learned edge message. Other ambiguities can persist when different internal routing choices produce the same observable evolution of the state. This situation can occur in networks containing cycles, insufficiently excited branches or a measurement set that does not distinguish the relevant branches. The intervention and threshold-sweep experiments provide evidence that the named gates are meaningful on the systems studied here. 

Finally, the real fluid-mixing experiment attempts at transferring architectural principles. Its held-out physical-fault results are satisfactory, but the leakage-plus-clogging and stirring-error categories each contain only one session. The tests performed on the algorithm suggest using CHGN as a structure-preserving simulator for systems in which the relevant state and external forcing are observed, while routing and operating mechanisms remain latent but physically defined and manipulable during prediction.

% ---------------------------------------------------------------------------
\section{Conclusions}
% ---------------------------------------------------------------------------

This work formulated learning on process networks whose effective routing is latent, state-dependent and dynamic. The Conservative Hybrid Graph Network predicts switch-controlled routing weights and local unit regimes, but constrains the ways in which these variables enter the state dynamics. The full-time evolution of the variables of interest is, in fact, reconstructed from both learned and fixed entries, thereby removing the direct gate-message rescaling pathway and guaranteeing that internal transport redistributes material without creating or destroying it.
The model was evaluated on a fixed-graph benchmark, where a less-constrained sequence model remained more accurate at point prediction and was extrapolated to larger graph sizes without retraining. A second benchmark involved a real fluid-mixing system, in which the architecture retained its forecasting ability for held-out physical faults that were not explicitly associated with manual operations.
The main result is therefore that CHGN provides routing variables that can be inspected and intervened on, an exact internal transport balance, and a single set of graph-local mappings that can be reused across process topologies.

\section*{Acknowledgements} The author gratefully acknowledges the support of King Abdullah University of Science and Technology (KAUST) through the Clean Energy Research Platform.

\bibliographystyle{unsrt}
\bibliography{references}

% ---------------------------------------------------------------------------
\appendix
% ---------------------------------------------------------------------------

\section{State reconstruction on the training topology}\label{app:reconstruction}

\begin{figure}[t]
\centering
\includegraphics[width=\textwidth]{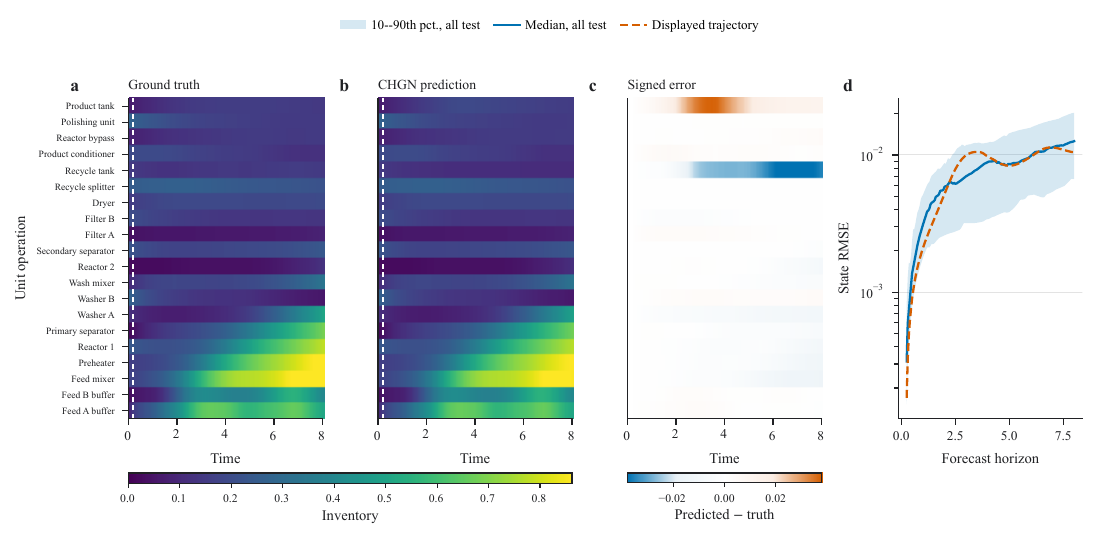}
\caption{\textbf{Full-trajectory state reconstruction.} (a) Ground-truth inventory for all twenty units; (b) CHGN rollout on the same scale; (c) signed residual on a scale two orders of magnitude finer. The dashed line marks the end of the conditioning history. (d) State RMSE versus forecast horizon: median and 10--90th percentile band over the test split, with the displayed trajectory dashed.}
\label{fig:state-reconstruction}
\end{figure}

Figure~\ref{fig:state-reconstruction} reports a complete autoregressive rollout over all twenty units. The model receives the initial history and is never teacher-forced afterwards. At the inventory scale, the predicted and true trajectories are almost indistinguishable. Panel~(c) therefore reports the signed residual on a scale two orders of magnitude finer. Most of the visible errors occur in the product and recycle tanks, where the biases have opposite signs and grow together. This behaviour is consistent with a small error at switch S5: material is sent slightly too often to one branch and slightly too little to the other. Because the transport update remains conservative, the routing error redistributes inventory between the two tanks instead of creating a common positive or negative drift. The horizon error rises during the first part of the forecast and then approaches a plateau near the aggregate rollout RMSE reported in Table~\ref{tab:baselines}.

\begin{figure}[t]
\centering
\includegraphics[width=\textwidth]{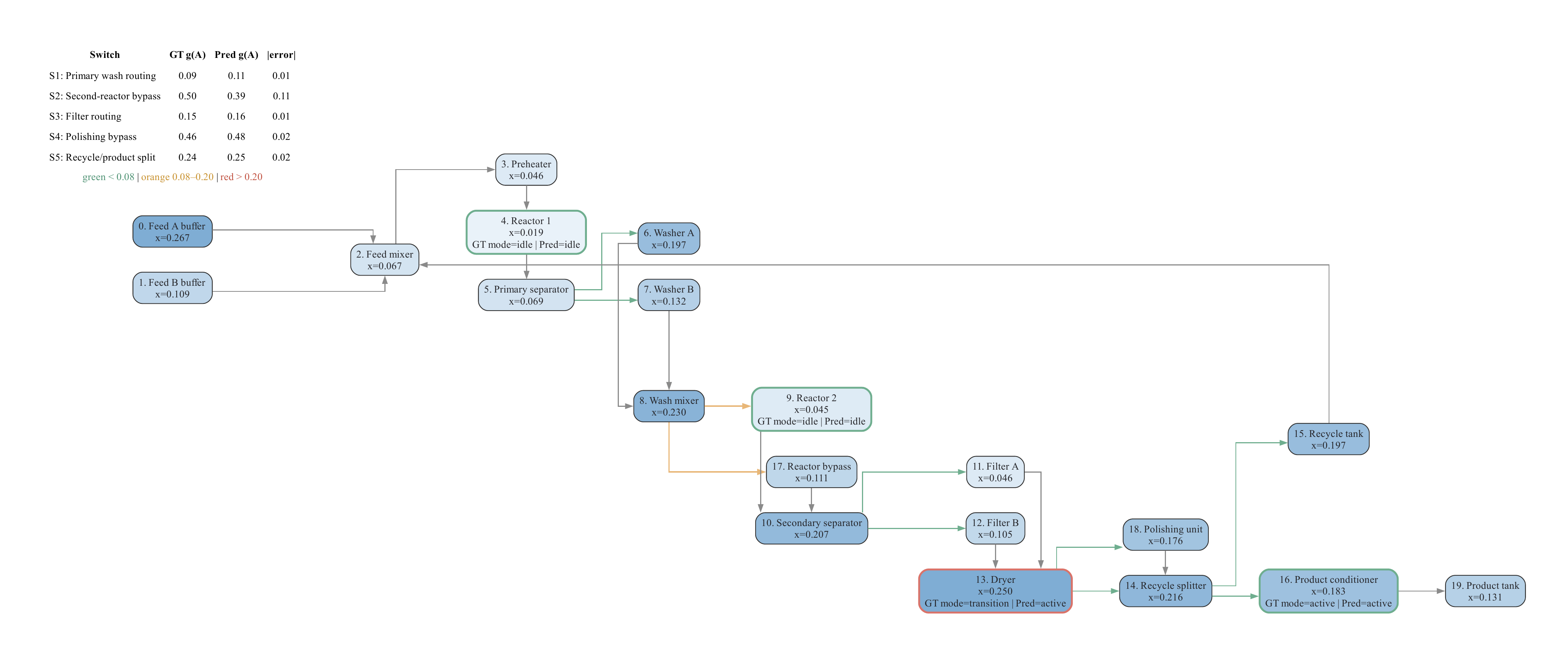}
\caption{\textbf{Recovered mechanism on the plant graph}, held-out trajectory 255 at $t=0.20$. Node labels show inventory and, for regime-bearing units, true and predicted regime; a red border marks disagreement. Edge colour shows absolute gate error (green $<0.08$, orange $0.08$--$0.20$, red $>0.20$), and the inset lists true and predicted gate values. Because all columns are rounded independently, the printed error at S1 and S5 differs by $0.01$ from the difference between the printed gate values. This instant was chosen to show the dryer and S2 errors; aggregate accuracy is reported in Table~\ref{tab:baselines}.}
\label{fig:flowsheet}
\end{figure}

Figure~\ref{fig:flowsheet} places the recovered mechanisms directly on the process graph at one held-out instant. Every unit reports its inventory and, where applicable, its true and predicted regime. Every switched branch reports the true gate, the predicted gate and the resulting error.

The selected instant contains two visible failures. The dryer is predicted to be active rather than transitioning, and the S2 bypass gate has an absolute error of $0.11$, compared with $0.01$--$0.02$ for the other switches. Both errors are consistent with the aggregate results. Transition is the least accurate regime class in Figure~\ref{fig:parity}c, while S2 lies close to $g=0.5$, where the logistic response of Equation~\eqref{eq:routing_sigmoid} is most sensitive to a threshold error because $g(1-g)$ is maximal at the centre of the sigmoid.

\section{Interventional audit and counterfactual response}\label{app:intervention}

\paragraph{One-step substitution.}
The learned and oracle mechanisms produce comparable one-step errors. The all-unit RMSE is $4.115\times10^{-4}$ with learned variables and $4.239\times10^{-4}$ when both mechanisms are replaced by their true values. True regimes alone leave the all-unit error unchanged to four significant digits. In contrast, fixing the routing increases the error by a factor of $4.7$ to $1.932\times10^{-3}$, while shuffling both mechanisms raises it to $1.987\times10^{-3}$. The effect also occurs where the mechanism acts. Fixed routing increases the error at switch destinations by a factor of $5.9$, while fixed regimes increase the error at regime-bearing units by a factor of $3.6$.

The learned variables are more accurate than the oracle substitution in two columns of Table~\ref{tab:intervention}.  The latter indicates that the learned gates can compensate for approximation error in the remaining parts of the operator. The evidence that a mechanism is load-bearing is therefore the error increase produced when it is fixed or shuffled, rather than the expectation that the oracle row must always be the numerical lower bound.

\begin{table}[t]
\centering
\small
\caption{Teacher-forced one-step intervention audit, $7488$ samples. Values are RMSE after substituting mechanism variables into the same trained operator; ratios are relative to the learned mechanism. \emph{Single seed}; see \S\ref{sec:limitations}.}
\label{tab:intervention}
\begin{tabular*}{\textwidth}{@{\extracolsep{\fill}}lccc@{}}
\toprule
Substitution & All units & Switch destinations & Regime-bearing units\\
\midrule
Learned                  & $4.115\times10^{-4}$ & $4.627\times10^{-4}$ & $1.084\times10^{-4}$\\
Oracle: both             & $4.239\times10^{-4}$ & $4.847\times10^{-4}$ & $0.778\times10^{-4}$\\
Oracle: routing only     & $4.240\times10^{-4}$ & $4.847\times10^{-4}$ & $0.786\times10^{-4}$\\
Oracle: regimes only     & $4.115\times10^{-4}$ & $4.627\times10^{-4}$ & $1.079\times10^{-4}$\\
\midrule
Fixed: regimes only      & $4.441\times10^{-4}$ & $4.622\times10^{-4}$ & $3.890\times10^{-4}$\\
Fixed: routing only      & $1.932\times10^{-3}$ & $2.710\times10^{-3}$ & $9.620\times10^{-4}$\\
Fixed: both              & $1.939\times10^{-3}$ & $2.709\times10^{-3}$ & $1.026\times10^{-3}$\\
Shuffled: both           & $1.987\times10^{-3}$ & $2.777\times10^{-3}$ & $1.358\times10^{-3}$\\
\midrule
\emph{Worst ablation / learned} & $4.8\times$ & $6.0\times$ & $12.5\times$\\
\bottomrule
\end{tabular*}
\end{table}

\paragraph{Fixed-history threshold sweeps.}
In the counterfactual audit, each threshold $\theta_s^g$ is swept while the held-out history remains fixed. The recovered response follows the true logistic response over the complete range. All five switch responses and four regime responses are monotonically decreasing over the entire sweep, yielding a monotone fraction of $1.000$. The switch-gate MAE ranges from $9.2\times10^{-3}$ to $2.1\times10^{-2}$, while regime agreement ranges from $96.4\%$ to $99.7\%$. S2 shows the only clear deviation, with an offset of approximately $0.04$ at the low-threshold end while remaining almost parallel to the true curve. The latter is consistent with a threshold bias rather than with failure to recover the response form.

\paragraph{Closed-loop, multi-step intervention effects.}
The one-step substitutions and fixed-history sweeps do not propagate the effect of an intervention through the future state. A third audit, therefore, changes one threshold by $\Delta=\pm0.08$ for switches or $\Delta=\pm0.04$ for regimes. For each of $24$ held-out trajectories, the simulator and CHGN then roll forward for $40$ steps without teacher forcing. The comparison is made through the intervention effect, defined as the counterfactual trajectory minus the factual trajectory for the same initial history.

Across $240$ routing interventions, the intervention-effect RMSE is $6.4\times10^{-4}$ and reaches $1.0\times10^{-3}$ at the final step. Across $192$ regime interventions, the corresponding RMSE is $8.0\times10^{-5}$, eight times smaller, and reaches $1.2\times10^{-4}$ at the final step. The absolute counterfactual-state error is nevertheless similar for the two groups, $4.05\times10^{-3}$ for routing and $3.99\times10^{-3}$ for regimes. The difference, therefore, reflects the smaller dynamical effect of the regime perturbations rather than a uniformly easier counterfactual prediction problem.

\section{Additional zero-shot transfer results}\label{app:transfer}

\begin{figure}[t]
\centering
\includegraphics[width=\textwidth]{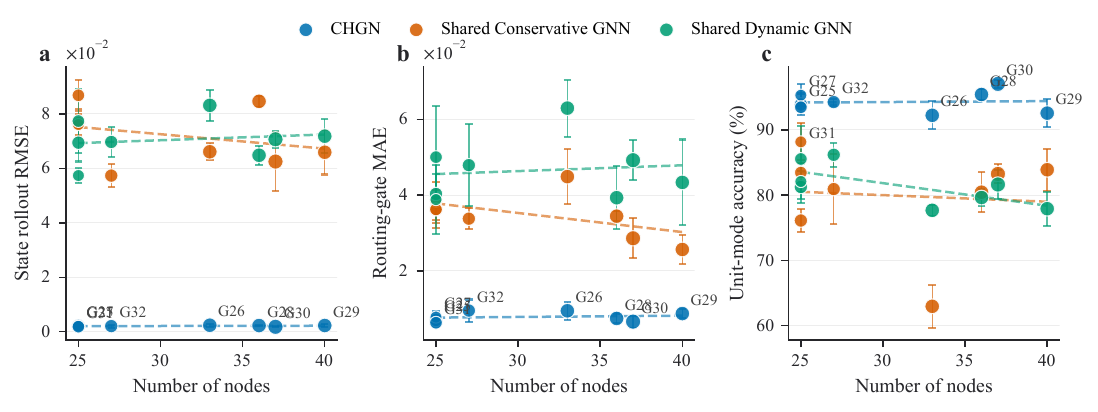}
\caption{\textbf{Zero-shot transfer as a function of graph size.} State rollout RMSE, routing-gate MAE and regime accuracy for CHGN, the Shared Conservative GNN and the Shared Dynamic GNN on unseen $25$--$40$-node process graphs. Dashed lines are descriptive least-squares trends. CHGN remains substantially more accurate across the full range of graph sizes.}
\label{fig:transfer-scaling}
\end{figure}

\begin{figure}[t]
\centering
\includegraphics[width=\textwidth]{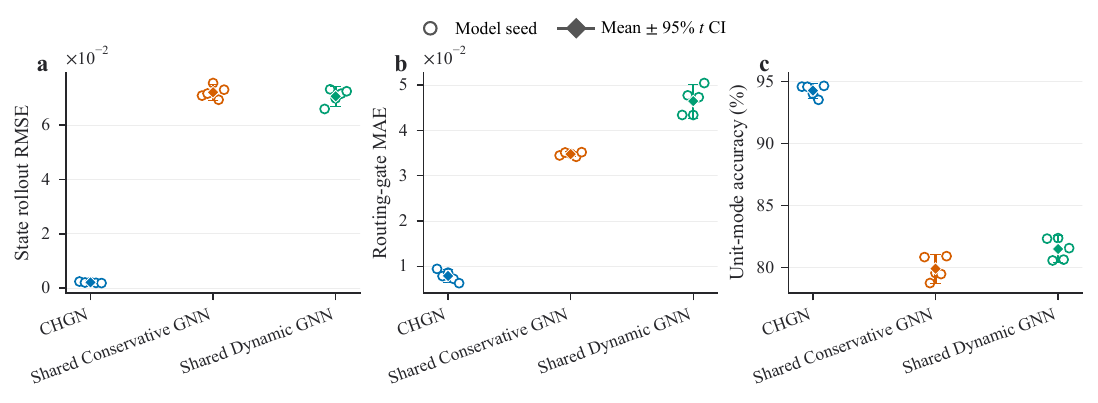}
\caption{\textbf{Seed robustness of zero-shot transfer.} Circles show independently trained model seeds; diamonds and bars show the mean and $95\%$ Student-$t$ interval. CHGN retains a lower state-rollout RMSE and routing-gate MAE, and higher regime accuracy, than both shared-weight baselines across seeds.}
\label{fig:transfer-seeds}
\end{figure}

Figures~\ref{fig:transfer-scaling} and~\ref{fig:transfer-seeds} resolve the transfer comparison of Figure~\ref{fig:transfer} by graph size and by training seed, respectively. The separation persists across the full $25$--$40$-node range and across independently trained seeds, so it is not due to a small number of favourable topologies or to a single fortunate initialisation.

Table~\ref{tab:graph-transfer} retains an earlier graph-specific comparison. The listed baselines are retrained separately on each held-out graph, thereby testing graph-specific fitting rather than zero-shot transfer. They are not a matched replacement for the shared-weight comparison of \S\ref{sec:transfer}. Under this secondary protocol, the best retrained models reach approximately $0.0305$ state RMSE, compared with $2.118\times10^{-3}$ for CHGN without parameter updates.

\begin{table}[t]
\centering
\small
\caption{Secondary graph-specific reference. CHGN is evaluated zero-shot on the eight held-out topologies (mean $\pm$ s.d.\ across five seeds, each averaged over all graphs). In contrast, the listed comparators are retrained from scratch on each held-out graph (mean $\pm$ s.d.\ across eight single-seed fits). Bold marks the zero-shot reference row; values in the two blocks are not directly comparable. These values complement, but do not replace, the matched shared-weight zero-shot comparison of \S\ref{sec:transfer}.}
\label{tab:graph-transfer}
\begin{tabular*}{\textwidth}{@{\extracolsep{\fill}}lccc@{}}
\toprule
Method & State RMSE $\downarrow$ & Gate MAE $\downarrow$ & Regime acc. $\uparrow$ \\
\midrule
CHGN (zero-shot) & $\mathbf{(2.118 \pm 0.230)\times 10^{-3}}$ & $\mathbf{(7.861 \pm 1.218)\times 10^{-3}}$ & $\mathbf{94.27 \pm 0.49\%}$ \\
\midrule
\multicolumn{4}{l}{\emph{Retrained per graph}} \\
MLP           & $0.07483 \pm 0.00712$ & $0.12386 \pm 0.01117$ & $55.71 \pm 9.13\%$ \\
LSTM          & $0.05374 \pm 0.00951$ & $0.09535 \pm 0.01492$ & $59.82 \pm 7.15\%$ \\
Fixed GNN     & $0.04395 \pm 0.01769$ & $0.32299 \pm 0.02731$ & $92.84 \pm 2.13\%$ \\
Dynamic GNN   & $0.03051 \pm 0.00924$ & $0.02101 \pm 0.00832$ & $95.62 \pm 1.14\%$ \\
NRI-inspired  & $0.03750 \pm 0.01092$ & $0.07431 \pm 0.01811$ & $93.55 \pm 2.12\%$ \\
NHA-inspired  & $0.03046 \pm 0.00924$ & $0.02086 \pm 0.00797$ & $95.28 \pm 1.78\%$ \\
\bottomrule
\end{tabular*}
\end{table}

\section{Fixed-graph predictive benchmark}\label{app:baselines}

\begin{figure}[t]
\centering
\includegraphics[width=\textwidth]{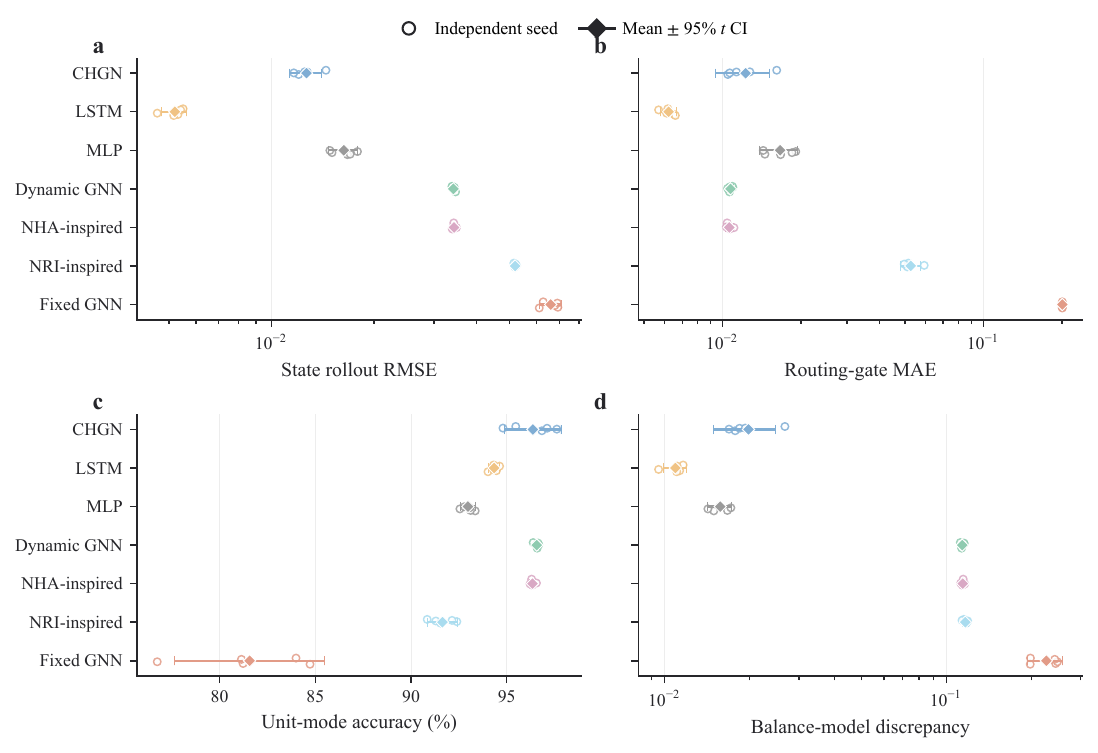}
\caption{\textbf{Fixed-graph predictive benchmark.} Circles show independent seeds; diamonds and bars show the mean and $95\%$ Student-$t$ interval. Error metrics use logarithmic axes. On this fixed graph, the LSTM leads panels (a), (b), and (d).}
\label{fig:baselines}
\end{figure}

Figure~\ref{fig:baselines} shows the seed-level distribution behind Table~\ref{tab:baselines}. The ordering of the models is stable across seeds, and a single outlying run does not produce the LSTM advantage on this fixed topology.

\section{Numerical physics audit}\label{app:numerics}

The numerical audit separates approximation error from structural consistency. Across independently trained seeds, the state-derivative RMSE is approximately $9.8\times10^{-3}$, indicating that finite-model form and discretisation errors persist in the learned dynamics. Reconstructing the reported state update directly from the predicted derivative yields zero RMSE for every seed, confirming that the stored update and the implemented Euler step are numerically consistent. The maximum internal-transport residual,

\begin{equation}
\max_\tau\left|\one^\top B\hat{\mathbf F}_\tau\right|,
\label{eq:transport_residual}
\end{equation}
is approximately $1.2\times10^{-7}$ and is therefore at the level expected from floating-point accumulation. The non-negativity correction in Equation~\eqref{eq:explicit_euler_loop} is also zero in RMS over every entry and rollout step.

The remaining derivative error reflects the approximation of the non-transport dynamics, as well as the difference between the Euler update used by CHGN and the RK4 integration used to generate the synthetic trajectories. The baselines do not expose a separate transport term, so the same structural audit cannot be applied to them. This absence of an auditable term does not by itself prove that a baseline violates conservation on every trajectory.

\end{document}